\documentclass[10pt,twocolumn,letterpaper]{article}
\usepackage[pagenumbers]{cvpr}
\usepackage{graphicx}
\usepackage{amsmath}
\usepackage{amssymb}
\usepackage{booktabs}
\usepackage[accsupp]{axessibility}

\usepackage[dvipsnames]{xcolor}

\definecolor{cvprblue}{rgb}{0.21,0.49,0.74}
\usepackage[pagebackref,breaklinks,colorlinks,citecolor=cvprblue]{hyperref}

\def\confName{CVPR}
\def\confYear{2024}

\usepackage{mathrsfs}
\usepackage{algorithm}
\usepackage{algpseudocode}
\usepackage{color}

\begin{document}

\title{A Lightweight Spatiotemporal Network for Online Eye Tracking with Event Camera}

\author{Yan Ru Pei\and
Sasskia Brüers\and
Sébastien Crouzet\and
Douglas McLelland\and
Olivier Coenen\\
{\tt\small \{ypei, sbruers, scrouzet, dmclelland, ocoenen\}@brainchip.com}\\
Brainchip Inc.\\
23041 Avenida De La Carlota, Suite 250, Laguna Hills, CA 92653
}

\maketitle

\begin{abstract}
Event-based data are commonly encountered in edge computing environments where efficiency and low latency are critical. To interface with such data and leverage their rich temporal features, we propose a causal spatiotemporal convolutional network. This solution targets efficient implementation on edge-appropriate hardware with limited resources in three ways: 1) deliberately targets a simple architecture and set of operations (convolutions, ReLU activations) 2) can be configured to perform online inference efficiently via buffering of layer outputs 3) can achieve more than 90\% activation sparsity through regularization during training, enabling very significant efficiency gains on event-based processors. In addition, we propose a general affine augmentation strategy acting directly on the events, which alleviates the problem of dataset scarcity for event-based systems. We apply our model on the AIS 2024 event-based eye tracking challenge, reaching a score of 0.9916 p10 accuracy on the Kaggle private testset.
\end{abstract}

\section{Introduction}

Event cameras are sensors that generate outputs (events) responding to optical changes in the scene's luminance \cite{event_vision}. For each pixel, an event is only produced when a brightness change is detected, which enables high levels of sparsity and temporal resolution when compared to traditional frame-based cameras. Given its high temporal resolution, often at sub-millisecond scales, data from event cameras contain very rich temporal features capturing subtle movement patterns.

However, when processing event data with modern neural networks, usually the events are pre-processed in ways that discard a large amount of temporal information. For example, oftentimes a spatial CNN is used to interface with event-based datasets, which requires the event data be ``collapsed" or ``binned" into images \cite{event_binning}. In some sense, this defeats the purpose of using event cameras in the first place, where the goal is precisely to capture these temporal subtleties. Furthermore, the process of binning into frames introduces an inevitable compromise in the temporal domain: too short, and the frames may not contain adequate event information; too long, and the low-latency advantage of the event-based data is lost\footnote{To minimize latency, one can apply the network on overlapping windows of the event stream, but this will greatly increase the computational requirement of the pipeline.}. 

In short, spatial CNNs are not efficient for either capturing temporal correlations of events or performing online inference efficiently on streaming data, as they inherently lack temporal continuity in the network. One way to compensate for this is to add a recurrent head (e.g. ConvLSTM) to the spatial CNN backbone \cite{conv_lstm}, in order to model temporal correlations by maintaining internal memories of high-level features. However, this temporal modeling only happens deep into the network, and far from the event inputs where low-level temporal details are prominent. Furthermore, this suffers from the typical difficulty of training recurrent networks in general \cite{bptt}.

Here, we propose a spatiotemporal convolutional network that efficiently performs online inference on streaming data, applying it on the challenge of event-based eye tracking. We make several contributions in our work. First, we design a light-weight spatiotemporal neural network that is fully causal, enabling it to perform online inference efficiently using FIFO buffers in place of storing all temporal frames. Second, we propose a causal event binning strategy that minimizes latency and reduces excessive buffering of event streams during inference. Third, we show that simply by including $L_1$ regularization on activations during training, sparsity (zero-valued outputs) per layer can be increased to greater than 90\%, enabling highly efficient inference on processors capable of exploiting that. Finally, we use a normalization strategy which alternates BatchNorm and GroupNorm layers \cite{group_norm} in the network, while fully preserving causality during inference.

\begin{figure*}[htbp]
  \centering
  \includegraphics[width=\textwidth]{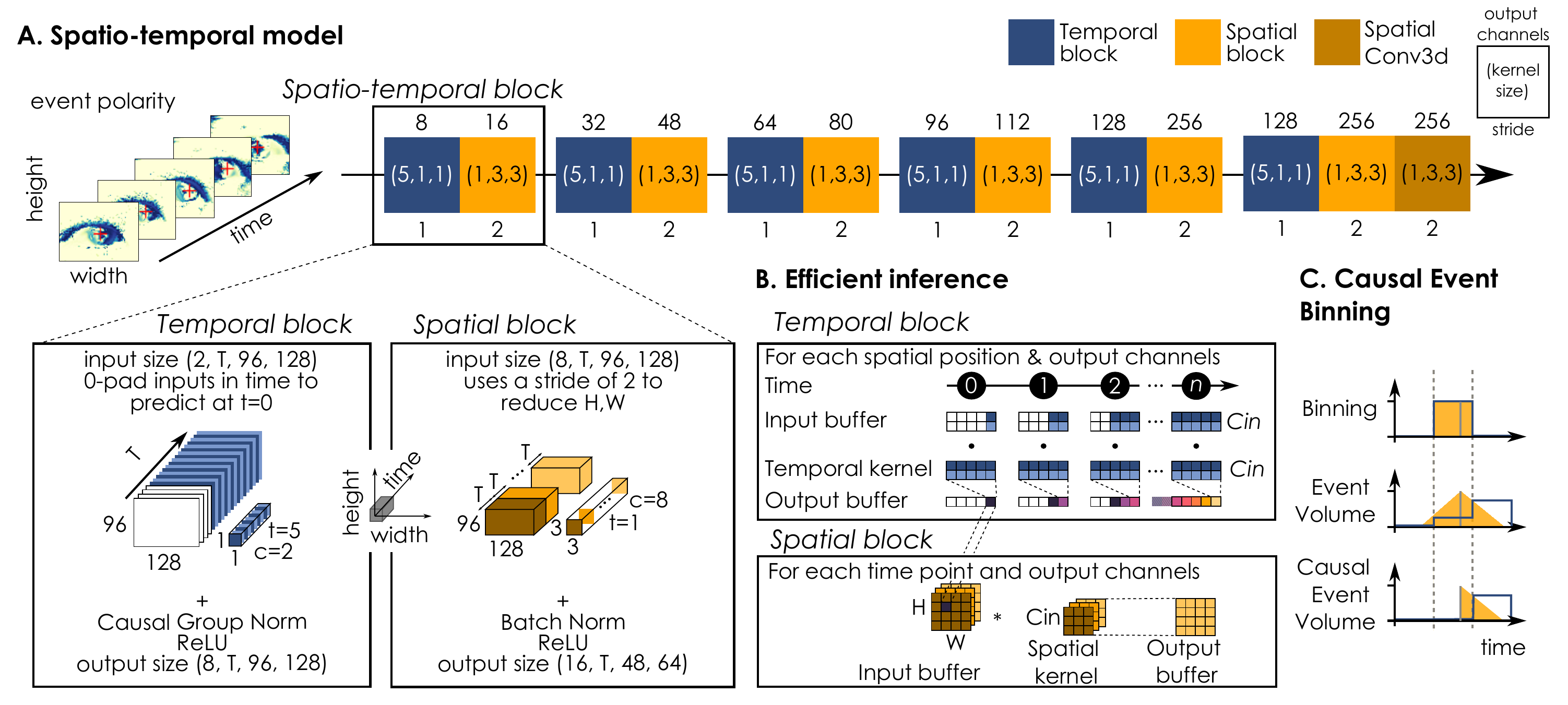}
  \caption{\textbf{A.} A lightweight spatiotemporal architecture for efficient eye tracking. The backbone is composed of a succession of 5 spatiotemporal blocks. Each spatiotemporal block consists of a temporal convolution followed by a spatial convolution. \textbf{B.} The model can be configured to run in streaming inference mode by using an input FIFO buffer for each temporal layer. The sliding-window mechanism of the FIFO buffer would act as the convolution sliding window, and the convolution operation itself is simply replaced by a dot product between the elements in the FIFO buffer and kernel weights. \textbf{C.} Compares the methods of direct binning, event volume binning, and causal event volume binning. The last method retains temporal information while still being fully causal.}
  \label{fig_network}
\end{figure*}

\section{Related Work}

\subsection{Event binning methods}

To feed event-based data into conventional neural networks, it is often necessary to bin them into uniform grids, so they can be passed as tensors to the networks. Many different event binning methods have been explored in the past, and it is beyond the scope of this article to provide a full taxonomy of these methods. Roughly, these methods can be categorized by the shape of the resulting tensor dimensions. 

Certain methods aggregate both polarity and temporal information, resulting in a single event frame shaped $(H, W)$ \cite{event_binning}. Other methods choose to preserve polarity information, while encoding the temporal information as either timestamps or count rates \cite{aes, hats}, resulting in a feature shaped $(2, H, W)$. Additionally, one can choose to preserve temporal frames while merging polarity information, resulting in event frames shaped $(T, H, W)$ \cite{event_volume, hots}. Finally, we have the most general case where the full event tensor of shape $(2, T, H, W)$ is kept \cite{est}. 

Among these methods, it is also common to apply a filtering or smearing pre-processing step on the raw events with either a fixed \cite{event_volume} or a trainable temporal kernel \cite{est} before binning. This allows the retention of finer temporal features of the events, often with added computational costs. In our work, we use a simple event volume smearing method that is causal and requires very little processing.

\subsection{Spatiotemporal networks}

There are several classes of neural networks with the capability to process spatiotemporal data (e.g. mostly videos). One such class of networks combines components from spatial convolutional networks and recurrent networks, such as the ConvLSTM model \cite{conv_lstm}. These types of models interface well with streaming spatiotemporal data, but is oftentimes difficult to train (as with recurrent networks in general). On the other hand, we have a class of easily trained networks that perform (separable) spatiotemporal convolutions \cite{r21d, p3d}, but are difficult to configure efficiently for online inference on streaming data.

In this work, we propose a general spatiotemporal convolutional network, where the temporal convolutional layers are all made causal. This network can be trained efficiently, and also perform online inference with minimal latency.

\subsection{Lightweight detector heads}

Object detection has become a standard task in computer vision, and many detector heads are developed for this purpose, most prominently the series of YOLO models \cite{yolo}. The eye-tracking task of this challenge can be considered a simpler version of the object detection task, where only the object's center is to be detected, with no bounding box dimensions to consider. To solve this task, we borrow components from modern detector heads \cite{center_net}: the network's output feature grid predicts at each point a binary logit of ``pupilness" along with spatial offsets of the pupil center (if present) relative to the grid point. Here we use standard classification and regression loss functions for object detection: a weighted combination of a focal loss with a smooth $L_1$ loss \cite{focal_loss, loss_functions}.

\section{Event data processing}

An event can be succinctly represented as a tuple $E = (p, x, y, t)$, where $p$ denotes the polarity, $x$ and $y$ are the horizontal and vertical pixel coordinates, and $t$ is the time. A collection of events can then be expressed as $\mathcal{E} = \{E_1, E_2, ..., E_n\}$, where $n$ is the number of events. In practice, events generated by event cameras differ from conventional RGB pixels in the following ways: 1) the polarity $p \in \{-1, +1\}$ typically contains only positive and negative information; 2) the temporal resolution of $t$ is much finer, usually at $1\text{ms}$ resolution; 3) the data is much sparser due to its high temporal resolution and dependence on motion.

\subsection{Event volume binning}

Various methods of binning have been studied, and we focus here on a binning procedure which converts events into a tensor shaped $(2, T, H, W)$ to preserve richer information from the raw event data than direct binning. One such method is ``event volume binning" \cite{event_volume}. One can visualize this process by first expanding an event (from a point) to an \textit{event volume} with a cuboid of the same dimensions as the bins. Typically, unless an event is centered perfectly in the bin, its event volume will span multiple adjacent bins. The spatiotemporal information is then encoded by the event volume's contributions in the bins it occupies.

More formally, we consider a bin at location $(x_b, y_b, t_b)$ and size $(\Delta x_b, \Delta y_b, \Delta t_b)$, and the set of events $\mathcal{E} = \{E_1, E_2, ..., E_n\}$, where each event is represented by the tuple $E_i = (p_i, x_i, y_i, t_i)$. The total amount of ``occupancy contribution" that the bin receives for each polarity is then
\begin{equation*}
\begin{split}
V_{+} &= \sum_{\{E_i \mid p_i = 1\}} k(\frac{x_b - x_i}{\Delta x_b}) k(\frac{y_b - y_i}{\Delta y_b}) k(\frac{t_b - t_i}{\Delta t_b}) \\
V_{-} &= \sum_{\{E_i \mid p_i = -1\}} k(\frac{x_b - x_i}{\Delta x_b}) k(\frac{y_b - y_i}{\Delta y_b}) k(\frac{t_b - t_i}{\Delta t_b}), \\
\text{where} \\
k(\chi) &= \max(| 1 - \chi |, 0)
\end{split}
\end{equation*}
is a triangle filter. Note that this allows us to perform temporal binning and spatial downsampling (if needed) all at once, and can be considered a version of bilinear interpolation for events.

\subsubsection{Causal Event volume binning}
\label{causal_event_volume}

For causal event volume binning, we simply modify the triangle filter acting in the temporal dimension into $k(\tau) = H(\tau) \max(|1 - \tau|, 0)$, where the added $H$ is the Heavide side step function:
\begin{equation*}
H(x) = 
\begin{cases} 
0 & \text{if } x < 0\,, \\
1 & \text{if } x \geq 0\,.
\end{cases}
\end{equation*}
In other words, while the spatial filter shape remains unchanged, the temporal triangle filter now becomes a half-triangle filter that only extends in the ``future" direction, making it a causal filter (see Fig ~\ref{fig_network} section C for a visualization of the temporal filter shape).

This means that to compute the volume contribution to the bin at time $t_b$, we no longer need to ``wait" for any future events where $t_i > t_b$, as they won't contribute to the bin. This makes the event pre-processing step in the inference pipeline introduce less intrinsic latency, as the bin can stream its value to the network as soon as all the events up to time $t_b$ are received.

\subsection{Spatial affine transformation of events}

Usually in computer vision, affine transformations are discussed in the context of data augmentations for images. However, they can easily be applied to event data using a fairly standard procedure. Working in homogeneous coordinates, the spatial location of an event is expressed as $(x', y', 1)$, with $x' = \frac{x}{W} - \frac{1}{2}$ being the relative $x$-offset from the center of the frame (similarly for $y'$). 

To apply the affine transformation to the events, we set up the scaling, rotation, and translation matrices and multiply them together as such:
\begin{equation*}
\begin{split}
    A &= TRS \\
    &=
    \begin{bmatrix} 1 & 0 & t_x \\ 0 & 1 & t_y \\ 0 & 0 & 1 \end{bmatrix}
    \begin{bmatrix} \cos(\theta) & -\sin(\theta) & 0 \\
    \sin(\theta) & \cos(\theta) & 0 \\ 0 & 0 & 1 \end{bmatrix}
    \begin{bmatrix} s_x & 0 & 1 \\ 0 & s_y & 0 \\ 0 & 0 & 1 \end{bmatrix},
\end{split}
\end{equation*}
where the order of transformations follows the \texttt{torchvision} convention. After pre-computing the affine matrix $A$, we then simply multiply the spatial coordinates of all events with this matrix, which is an operation easy to parallelize
\begin{equation*}
(TRS)
\begin{bmatrix}
x'_1 & y'_1 & 1 \\ x'_2 & y'_2 & 1 \\ ... \\ x'_n & y'_n & 1
\end{bmatrix}.
\end{equation*}
After performing this transformation, we remove the last column of ones, and convert the relative center offsets back into pixel coordinates. Note that the pupil center labels need to be transformed along with the event data, which can be done using the same affine matrix.

In our study, we apply a random affine transformation to each event segment (but do not vary the affine parameters within the segment). We only apply part of the operations usually included in affine transformation: we apply scaling, rotation and translation and don't consider the reflection or shear transformation. The scaling factors $s_x$ and $s_y$ are uniformly and independently sampled between 0.8 and 1.2; the angle of rotation is uniformly sampled between -15 and 15 degrees; the two translation factors $t_x$ and $t_y$ are uniformly sampled between -0.2 and 0.2 independently.

Due to our affine augmentation strategy, the pupil location may be out-of-bounds in certain frames, so we ignore these frames. We also ignore frames where the eye is closed as labeled.

\subsection{Temporal affine transformation of events}
\label{temporal_affine}

Similarly to the spatial affine transformations, we also use temporal affine transformations, in the form of temporal scaling and shifting. 

Before performing the binning procedure, we transform the timestamps of the events as $at + b$, where $a$ is a random scaling factor and $b$ is a shift factor\footnote{After this transformation, the label timestamps need to be realigned with the binning ones: we use linear interpolation to infer the new label coordinates at the binning timestamps.}. In this study, we chose $a$ to be a random number uniformly sampled between 0.8 and 1.2, and $b$ to be simply 0, meaning that we are not performing temporal shifts\footnote{The reason for this is because both our event volume binning strategy and the temporal convolution layers in the network are already time-translation invariant}.


After binning, we also randomly choose to invert the temporal dimension (along with the polarity) of the event frames, with $0.5$ probability. Note that ``playing backward" an event stream requires swapping the positive and negative polarities. More formally, if we consider the brightness of a pixel to be a function of time, then the first derivative of the function is flipped if the domain of the function (time) is flipped.

\section{Network Architecture}

Very briefly, the network architecture can be considered as a spatiotemporal CNN attached with a detection head similar to CenterNet \cite{center_net}. However, there are several key innovations, particularly for ensuring causality and lightweightness of the network that we will detail in this section. See Fig.~\ref{fig_network}.A for the network architecture used in this work. The official implementation of our network is in \cite{tenn_eye_code}.

\subsection{Causal spatiotemporal convolutional block}

The backbone of our network is a stack of spatiotemporal convolution blocks, with each block consisting of a temporal convolution layer followed by a spatial convolution layer. During training, the input to this block is a spatiotemporal feature tensor of shape (channels, frames, height, width). The convolution with the temporal kernel is performed on each spatial pixel independently, or equivalently a 3D convolution with kernel size of $k_t \times 1 \times 1$. Similarly, the spatial kernel is then performed on each feature frame independently, or equivalently a 3D convolution with kernel size of $1 \times k_x \times k_y$. 

The idea of performing this separation of spatial and temporal convolutions, instead of doing an expensive 3D convolution with a kernel sized $k_t \times k_x \times k_y$ has been explored in the literature, and is referred to as a pseudo-3D or (2+1)D convolution \cite{r21d, p3d}. However, we make several additional innovations on top of this structure to make the network more suitable to perform online inference.

First and foremost, we pre-pad our inputs by $k_t - 1$, which makes the temporal convolution performed in our block fully causal: the current output feature frame only depends on the current and previous input feature frames. This guarantees that the network is able to make predictions in real time, without delay. On the other hand, if we consider a non-causal centered temporal convolution with kernel size of 3, then each temporal convolution layer needs to ``wait" for a future frame to generate an output, and each such layer in the network will incur one additional frame of delay. The causality of the block allows us to choose a longer than typical temporal kernel size of $k_t = 5$ without costs to real-time inference: the generation of the current output feature only requires the past four plus current input frames. Using a longer temporal kernel allows us to leverage the higher input frame rate to our network.

Secondly, we choose to perform the temporal convolutions before the spatial convolutions to position the first temporal layer to have direct access to the pre-processed event data. Because event-based data contains potentially sensitive and useful temporal features, performing a temporal convolution on them is conducive to capturing more effectively these temporal features. Performing a spatial convolution first may ``smear" away these features\footnote{In some sense, the first temporal layer can be viewed as an extra layer of adaptive temporal filtering in addition to the fixed causal triangle filter we used before binning (see Section~\ref{causal_event_volume}).}. This makes our spatiotemporal block effectively an (1+2)D convolution.

Thirdly, unlike the R(2+1)D network, we opt not to use any residual connections as they can be potentially expensive when implemented on mobile hardwares, since buffering the input for the residual connection incurs extra memory requirement \cite{skip_fifo}. This cost is exacerbated if the network is pipelined\footnote{For a pipelined network interfacing with streaming data, a FIFO buffer of depth equal to the number of layers in the main path needs to be maintained in the skip path, in order to synchronize with the output from the main path.}.

Finally, each of the temporal and spatial layers can also be configured to be depthwise separable, to further reduce the compute requirement of the network. Depthwise separable spatial CNNs are amenable for deployment at low-resource environment \cite{mobile_net}, while suffering from little to no performance degradation. Here, we apply the same concept to a temporal convolution layer, decomposing it into a depthwise temporal convolutional layer followed by a pointwise convolutional layer.

\subsubsection{Causal group normalization}
\label{causal_gn}

In this work, we use a mixed normalization strategy and apply batch normalization after the spatial convolutional layers, and group normalization after the temporal convolutional layers. By doing so, we have the advantage of acquiring both fixed sample statistics during training (via BatchNorm), and the ability to perform dynamic normalization adapting to data during inference (via GroupNorm)\footnote{Note that even though the discussion here focused on group normalization, it also applies to other dynamic normalization methods such as instance and layer normalization.} \cite{group_norm}. This mixed normalization strategy allows for stability in both small and large batch-size training regimes (see Fig.~\ref{fig:bn}).

Since the batch normalization statistics are fixed and data independent during inference, we do not have to worry about its causality. However, for group normalization, which dynamically performs data-based normalization at inference time, ensuring causality is crucial: the dynamic normalization statistics need to be independent of future frames.

Recall that the event tensor is shaped $(C, T, H, W)$ during training, so the simplest way to ensure causality of in GroupNorm layers is to compute the statistics only over the last two spatial dimensions $(H, W)$. This way the statistics are never ``mixed" in time. This is the method we chose in this work. However, there are extensions of this approach that can also be considered. For instance, one can maintain a finite running window containing the past few frames for computing the normalization statistics. Alternatively, running statistics can be maintained which continually update based on the new feature frame\footnote{This is akin to BatchNorm except this statistics is updated temporally instead of through training batches.}.

\subsection{Configuration for online inference with FIFO buffers}

\begin{algorithm*}
\caption{Spatiotemporal Convolution Block (Inference mode with FIFO buffers)}
\label{st_block_inference}
\begin{algorithmic}
\State \textbf{Input:} $X \in \mathbb{R}^{C_{in} \times H \times W}$
\State \textbf{FIFO buffer (stateful):} $H \in \mathbb{R}^{5 \times C_{in} \times H \times W}$
\State \textbf{Parameters:}
\State \quad Channels: $C_{in}$, $C_{mid}$, $C_{out}$
\State \quad Kernel sizes: $K_{temp} = (C_{out}, C_{in}, 5)$, $K_{spatial} = (C_{out}, C_{in}, 3, 3)$
\State \quad Strides: $S_{spatial} = (1, 2, 2)$

\State
\State $H_{updated} \gets \Call{Concat}{H[1:], X[\text{None}, :]}$ \Comment{updates the FIFO buffer with new input data}
\State $X_{temp} \gets \text{einsum}(k_tc_{in}hw \,,\, c_{mid}c_{in}k_t \,\rightarrow\, c_{mid}hw,\, H_{updated},\, K_{temp})$ \Comment{Contracts temporal dimension}
\State $X_{gnorm} \leftarrow \text{GroupNorm}(X_{temp}, \text{groups}=4)$ \Comment{Applied to each feature frame separately (for causality)}
\State $X_{med} \leftarrow \text{ReLU}(X_{gnorm})$
\State
\State $X_{spatial} \leftarrow \text{Conv2D}(X_{med}, K_{spatial}, S_{spatial}, \text{padding}=\text{same}, C_{mid} \rightarrow C_{out})$
\State $X_{bnorm} \leftarrow \text{BatchNorm}(X_{spatial})$
\State $X_{output} \leftarrow \text{ReLU}(X_{bnorm})$
\State
\State \textbf{Output:} $X_{output} \in \mathbb{R}^{C_{out} \times \frac{H}{2} \times \frac{W}{2}}$
\end{algorithmic}
\end{algorithm*}

Since we are performing causal temporal convolutions in every temporal layer of the network, we can use a FIFO buffer as the input buffer to each temporal layer to handle streaming input features to the layer. See Algorithm~\ref{st_block_inference} for a pseudo-code of how a spatiotemporal block runs in online inference mode, and see Fig.~\ref{fig_network}.B for a visualization of the operations.

At a high level, a convolution operation (considering only 1 input and output channel) can be viewed as a dot product between a sliding window of the input and the temporal kernel. We can isolate the sliding window part of the operation, and let it be naturally performed by the FIFO buffer mechanism. The depth of the FIFO buffer is simply the temporal kernel size. As soon as the buffer is updated, the dot product operation can be performed to generate the output feature map. This is possible due to the causality of the network, where the buffer does not need to ``wait" for any future input features to generate the current output feature.

\subsection{The detector head and loss}

After the data passes through the backbone network, we get a feature map of size $C \times T \times 3 \times 4$, which is passed through an additional temporal layer for smoothing before feeding it to the detector head. Inspired by the simple CenterNet detector \cite{center_net}, our head consists of a $3 \times 3$ spatial convolution, a ReLU activation, then a $1 \times 1$ spatial convolution, and a sigmoid function applied to each temporal frame separately, the output of which is a prediction of size $3 \times T \times 3 \times 4$. As with typical detection predictions, this map can be considered a $3 \times 4$ grid overlaid on top of the event frames, each grid-cell in each frame containing a prediction of: 1) the probability of a pupil being inside the cell, and 2) the relative x and 3) y offset of the pupil in the cell.

 We apply the following loss function to each grid-cell
\begin{equation*}
\ell = 
\begin{cases} 
-(1 - \hat{p})^{\gamma} \log(\hat{p}) + \ell_{\text{reg}} & \text{if } p = 1 \\
\hat{p}^{\gamma} \log(1 - \hat{p}) & \text{if } p = 0
\end{cases},
\end{equation*}
where the focal loss parameter $\gamma=2$ and $\hat{p}$ is the predicted probability of pupil presence and $p$ is the ground truth presence.

The regression loss $\ell_{\text{reg}}$ is the summed SmoothL1Loss $\ell_{1, \text{smooth}}$ for the $\hat{x}$ and $\hat{y}$ offset predictions of the network in each grid cell. For example, the SmoothL1Loss between the $\hat{x}$ prediction and the $x$ ground truth is, 
\begin{equation*}
\ell_{1, \text{smooth}}(\hat{x}, x) 
=
\begin{cases} 
0.5 \times \frac{(\hat{x} - x)^2}{\beta} & \text{if } |\hat{x} - x| < \beta \\
|\hat{x} - x| - 0.5 \times \beta & \text{otherwise}
\end{cases},
\end{equation*}
where $\beta = 0.11$. 

The total loss is computed by averaging over all grid-cells and valid frames (where the eye is open and within bounds).

\subsection{Activity Regularization}
Data from event-based sensors can be extremely sparse spatially and temporally, so there is a tendency to think that they will be well suited to processors (typically ``neuromorphic") that can exploit that sparsity by not processing zero-valued inputs. However, with default training, sparsity in the inputs does not necessarily translate to sparsity in outputs from later layers, simply because the model has no incentive to be ``quiet". To explore the extent to which sparsity within the network could be increased without impacting accuracy of predictions, we added activity regularization to the loss ($L_1$ norm on the outputs from ReLU layers, normalized by the volume of the outputs, and scaled by a weighting factor in the global loss).

\section{Experiments}

\subsection{Data and training pipeline}

The AIS 2024 eye tracking challenge dataset contains 13 subjects, each with 2 to 6 recording sessions \cite{wang2024ais_event}. The data is captured with an event camera with resolution of $480 \times 640$. The ground truth labels are provided at 100Hz, containing the x and y pixel coordinates of the pupil, and whether the eye is closed. 

The evaluation on the test set via the Kaggle API is performed at 20Hz. The test metric is the percentage of correct predictions within 10 pixels of the ground truth labels, in the spatially downsampled space of $60 \times 80$. It only considers labels where the eye is opened.

For training, we use a batch size of 32, where each batch consists of 50 event frames. We train for 200 epochs with the AdamW optimizer with a base learning rate of 0.002 and a weight decay of 0.005. We use the cosine decay with linear warmup scheduler, with warmup steps equal to 0.025 of the total training steps. We use automatic mixed-16 precision along with PyTorch compilation during training.

\subsection{Results and Ablation study}
\label{ablation}

\begin{table}[htbp]
\centering
\caption{\textbf{Challenge results.} Test results are from the submitted benchmark model trained on the full dataset available (train+validation). Validation performance is from the benchmark model trained only on the training set.}
\label{table:results-main-table}
\begin{tabular}{@{}ll@{}}
\toprule
\textbf{Benchmark model} & \textbf{p10} \\
\midrule
On public test set      & 0.988   \\
On private test set     & 0.992   \\
On validation set       & 0.963    \\
\end{tabular}
\end{table}

Overall, our model achieved 0.9898 on the test set. The performance of our model on the validation set, the public test set, and the private test set are reported in Table~\ref{table:results-main-table}. For inference and model evaluation, we feed entire segments to the network without splitting because we want to eliminate the artifacts of implicit padding as our network is agnostic to the number of input frames. 

We analyzed the contributions to our final results of different components of the data augmentation policy (Table~\ref{table:ablation-data}) and modification of the event processing method or model architecture (Table~\ref{table:ablation-model}). We report both the accuracy, with a 10-pixel error allowed (i.e. p10), and the average distance in pixels from ground-truth. Note that the values reported in this section are calculated on the validation set as defined in the challenge's original code repository.

\begin{table}[htbp]
\centering
\caption{\textbf{Ablation study results for data augmentation methods} included in the training pipeline. Each ablation was repeated 10 times and the results presented here are the average.}
\label{table:ablation-data}
\begin{tabular}{@{}lll@{}}
\toprule
\textbf{Data augmentation} & \textbf{p10} & \textbf{distance}\\
\midrule
Benchmark                    & 0.963   & 2.79 \\
w/o Spatial affine augment   & 0.588   & 11.32 \\
w/o Temporal flip augment    & 0.955   & 3.03 \\
w/o Temporal scale augment   & 0.968   & 2.64 \\
\bottomrule
\end{tabular}
\end{table}

\begin{table}[htbp]
\centering
\caption{\textbf{Ablation study results for modifications of the model architecture and event processing methods.} The default parameters are highlighted in \textit{italics}. \textit{CenterNet with DWS temporal} means the temporal smoothing layer before the head is depthwise-separable. BN is Batch Normalization, and GN is Group Normalization. MACs is the number of multiply-accumulate operations done per event frame, not accounting for sparsity. Results are the average of 10 repeats.}
\label{table:ablation-model}
\begin{tabular}{@{}lllll@{}}
\toprule
 & \textbf{p10} & \textbf{dist.} & \textbf{params} & \textbf{MACs} \\
\midrule
\textit{Benchmark}           & 0.963 & 2.79 & 809K & 55.2M \\
                             &       &      \\
\multicolumn{5}{l}{\textbf{Event-processing} - \textit{Causal event volume}} \\
$\rightarrow$ Event volume   & 0.959 & 2.77  & - & - \\
$\rightarrow$ Binning        & 0.959 & 2.74  & - & - \\
                             &       &      \\      
\multicolumn{5}{l}{\textbf{Model head} - \textit{CenterNet with DWS temporal}} \\
$\rightarrow$ Full temporal  & 0.964 & 2.72 & 1.07M & 58.3M \\
$\rightarrow$ No detector    & 0.936 & 3.52 & 216K & 47.4M \\
                             &       &      \\
\multicolumn{5}{l}{\textbf{Normalization} - \textit{Mixed}} \\
$\rightarrow$ All BN  & 0.969 & 2.61 & - & - \\
$\rightarrow$ All GN  & 0.960 & 2.82 & - & - \\
                             &  & &  & \\
\multicolumn{5}{l}{\textbf{Temporal kernel size} - \textit{5}}   \\
$\rightarrow$ 3       & 0.955  & 3.20 & 801K & 46.9M \\
                      &  & \\
\multicolumn{5}{l}{\textbf{Spatiotemporal block} - \textit{(1+2)D}} \\
$\rightarrow$ Conv3D    & 0.969 & 2.50 & 1.21M & 267M \\

\bottomrule
\end{tabular}
\end{table}

From Table~\ref{table:ablation-data}, we see that performing spatial affine augmentation of the event data provided boosted the performance of our model by 0.375. For the temporal affine augmentations, flipping the frames also slightly improved the performance. Surprisingly, temporal scaling seemed to have negative effects on the model performance. This may be due to poor choice of the range of temporal scaling factors (see Section~\ref{temporal_affine}), where the pupil movement patterns became too unnatural.

\begin{figure}[htbp]
  \centering
  \includegraphics[width=0.6\linewidth]{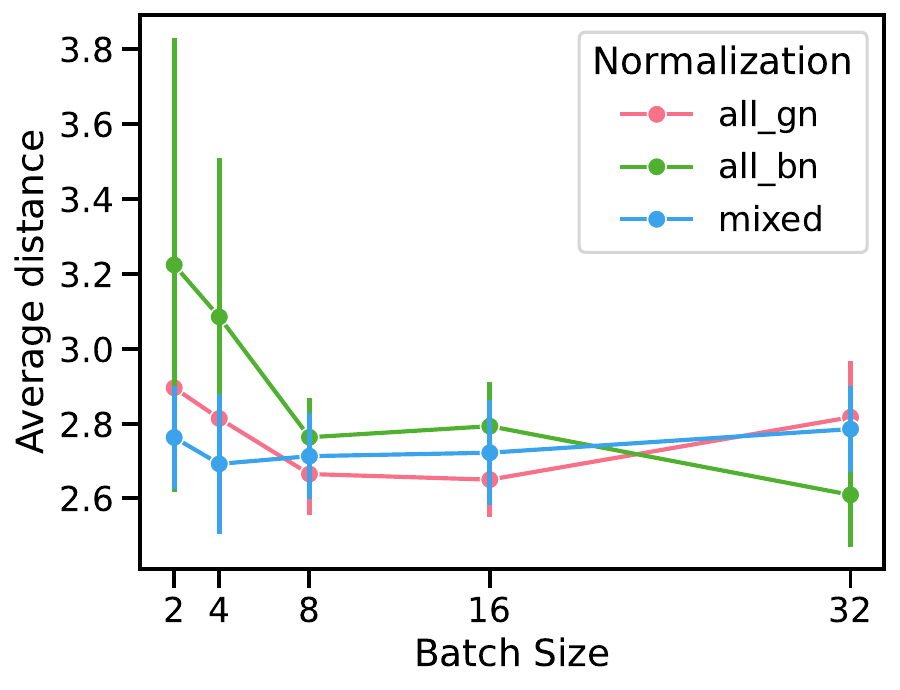}
  \caption{The average distance deviation with respect to batch size for networks using only causal group norms, only batch norms, and a mixture of causal group norms and batch norms.}
  \label{fig:bn}
\end{figure}

From Table~\ref{table:ablation-model}, we see that the event processing choices had very little impact on accuracy, but our benchmark (with causal event volume binning) ensures causality in the model as described in \ref{causal_event_volume}. For architectural choices, we see that using a CenterNet like head led to a boost of 0.027, compared with the ``no detector" (a simpler head with global average pooling and 2 dense layers). Similarly, using a larger temporal kernel led to a small boost in p10 accuracy (0.008) at a small computation cost (8.3M MACs). Note that while the model with only batch norms seems to perform better, this is highly dependent on the batch size of 32 used here (Fig.~\ref{fig:bn}). 

Fig.~\ref{fig:bn} shows the impact of normalization type on model performance depending on the batch size (as discussed in Section.~\ref{causal_gn}). We see the expected trend where batch norm performs better in the large-batch regime, and group norm performs better in the low-batch regime \cite{group_norm}. Using a mixture of batch norm and group norm seems to incorporate the best of both worlds, remaining relatively stable and high performing under batch-size variation\footnote{This is especially useful for training methods which requires dynamically annealing the batch size and sequence length (or number of frames).}.

\subsection{Efficiency-Accuracy Trade-offs}

For edge or low-power devices, accuracy is not the only criterion to consider when designing a model for a specific use case. The size to compute ratio is also of crucial importance. We tested three possible axes along which to optimize our model: input spatial resolution, decomposed convolutions, and increased activation sparsity.

\begin{figure}[htbp]
  \centering
  \includegraphics[width=\linewidth]{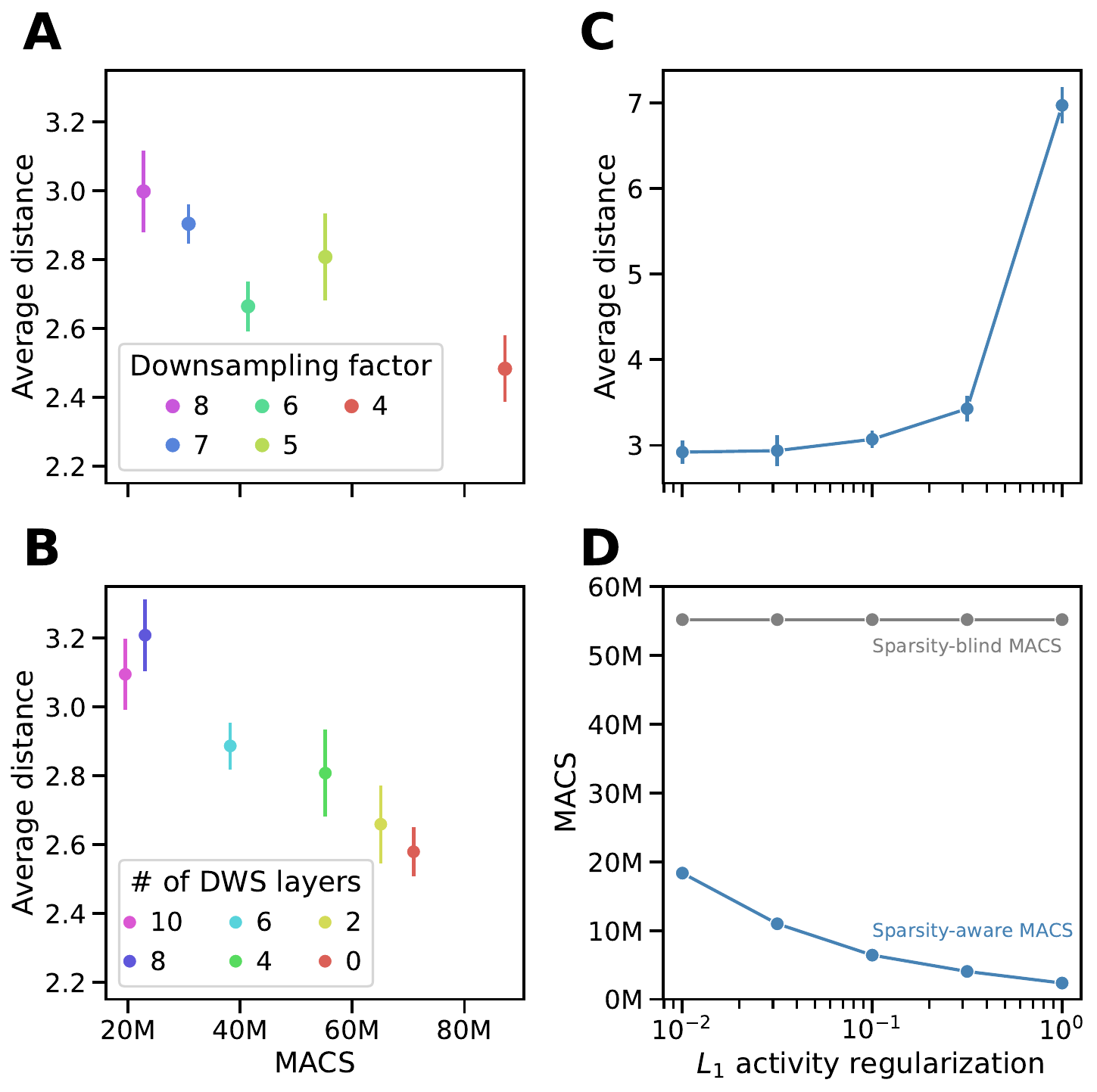}
  \caption{Default parameters in the benchmark network: spatial downsampling factor of 5, 6 depthwise-separable (DWS) layers, and no activity regularization. \textbf{A.} and \textbf{B.} The average distance deviation of the model predictions vs. the MACs per inference frame, as the input spatial downsampling factor and number of DWS layers are varied respectively. \textbf{C.} The average distance deviation vs. the weighting of the $L_1$ regularization loss. \textbf{D.} The temporally average MACs per inference frame vs. the regularization weighting for a sparsity-aware and a sparsity-blind processor.}
  \label{fig:optim}
\end{figure}

First, we tested the dependence of model performance on input spatial resolution. As shown in Figure~\ref{fig:optim}.A, even with the downsampling factor as high as 8 (inputs at 60 x 80), performance is only slightly worse than our default configuration (downsampling factor of 5), while dividing the compute requirements by 3.

Knowing the large compute gains that can be achieved via decomposed convolutions, from the outset we chose to use separate temporal and spatial convolution blocks in our architecture \cite{r21d}. Using full 3D convolutions can boost performance slightly (Table~\ref{table:ablation-model}) but at the cost of significantly increased memory and computational load. We can further decrease the computational load by making each temporal and spatial layer configurable as depthwise-separable (DWS) layers, as inspired by the MobileNet models \cite{mobile_net}. Fig.~\ref{fig:optim}.B shows the model performance as we increase the number of DWS layers, starting from the last layer of the backbone to the first layer\footnote{The computational savings of using DWS will be more pronounced in deeper layers where the number of channels generally increases.}.

We measured activation sparsity in the network (over the validation set), and found it to average around 50\%, about what one would expect given ReLU activation functions. Reasoning that much of this activity may not be informative, given the very high spatial sparsity of input to the network, we applied $L_1$ regularization to activation layers. Fig.~\ref{fig:optim}.C and \ref{fig:optim}.D shows the scaling of model performance and sparsity-aware MACs as the weighting of the regularization in the loss is varied. Extraordinarily, it is apparent that the model can achieve greater than 90\% sparsity while suffering minimal performance drop. On processors capable of fully exploiting sparsity, this directly translates to a 5x inference speedup vs. the default configuration. 

\section{Conclusion}

We introduced a causal spatiotemporal convolutional network designed to perform online inference efficiently. We also proposed a general spatiotemporal event affine augmentation strategy, and tested the efficacy of each of its components. Finally, we were able to drastically sparsify the activity of our network with little degradation in prediction precision. We believe that the current proposed network is fairly general and may be efficiently adapted to various domains with streaming spatiotemporal data, such as performing online multiclass object tracking with event or frame-based video streams. 


\raggedbottom
\pagebreak


\end{document}